\documentclass{article}

\usepackage{microtype}
\usepackage{graphicx}
\usepackage{subfigure}
\usepackage{booktabs} 

\usepackage{hyperref}

\usepackage[accepted]{icml2025}

\usepackage{amsmath}
\usepackage{amssymb}
\usepackage{mathtools}
\usepackage{amsthm}

\usepackage[capitalize,noabbrev]{cleveref}
\usepackage{multirow}

\theoremstyle{plain}

\theoremstyle{definition}

\theoremstyle{remark}

\usepackage[textsize=tiny]{todonotes}

\begin{document}

\twocolumn[
\icmltitle{Causal Machine Learning for Surgical Interventions}



\icmlsetsymbol{equal}{*}

\begin{icmlauthorlist}
\icmlauthor{J. Ben Tamo}{ece}
\icmlauthor{Nishant S. Chouhan}{ece}
\icmlauthor{Micky C. Nnamdi}{ece}
\icmlauthor{Yining Yuan}{cse}
\icmlauthor{Shreya S.  Chivilkar}{ece}
\icmlauthor{Wenqi Shi}{ut}
\icmlauthor{Steven W. Hwang}{shc}
\icmlauthor{B. Randall Brenn}{shc}
\icmlauthor{May D. Wang}{ece,bme}
\end{icmlauthorlist}

\icmlaffiliation{ece}{School of Electrical and Computer Engineering, Georgia Institute of Technology, Atlanta, USA}
\icmlaffiliation{cse}{School of Computational Science and Engineering, Georgia Institute of Technology, Atlanta, USA}
\icmlaffiliation{shc}{Shriners Children's, Philadelphia, Pennsylvania, USA}
\icmlaffiliation{ut}{School of Public Health, UT Southwestern Medical Center, Dallas, USA}
\icmlaffiliation{bme}{Wallace H. Coulter Department of Biomedical Engineering, Georgia Institute of Technology and Emory University, Atlanta, GA, USA}

\icmlcorrespondingauthor{J. Ben Tamo}{jtamo3@gatech.edu}
\icmlcorrespondingauthor{May D. Wang}{maywang@gatech.edu}

\icmlkeywords{causal inference, personalized healthcare, heterogeneous treatment effects, multi-task learning}

\vskip 0.3in
]



\printAffiliationsAndNotice{}  

\begin{abstract}
Surgical decision-making is complex and requires understanding causal relationships between patient characteristics, interventions, and outcomes. 
In high-stakes settings like spinal fusion or scoliosis correction, accurate estimation of individualized treatment effects (ITEs) remains limited due to the reliance on traditional statistical methods that struggle with complex, heterogeneous data.
In this study, we develop a multi-task meta-learning framework, X-MultiTask, for ITE estimation that models each surgical decision (e.g., anterior vs. posterior approach, surgery vs. no surgery) as a distinct task while learning shared representations across tasks. To strengthen causal validity, we incorporate the inverse probability weighting (IPW) into the training objective.
We evaluate our approach on two datasets: (1) a public spinal fusion dataset (1,017 patients) to assess the effect of anterior vs. posterior approaches on complication severity; and (2) a private AIS dataset (368 patients) to analyze the impact of posterior spinal fusion (PSF) vs. non-surgical management on patient-reported outcomes (PROs).
Our model achieves the highest average AUC (0.84) in the anterior group and maintains competitive performance in the posterior group (0.77). It outperforms baselines in treatment effect estimation with the lowest overall $\hat{\epsilon}_{\text{NN-PEHE}}$ (0.2778) and $\hat{\epsilon}_{\text{ATE}}$ (0.0763). Similarly, when predicting PROs in AIS, X-MultiTask consistently shows superior performance across all domains, with $\hat{\epsilon}_{\text{NN-PEHE}}$ = 0.2551 and $\hat{\epsilon}_{\text{ATE}}$ = 0.0902.
By providing robust, patient-specific causal estimates, X-MultiTask offers a powerful tool to advance personalized surgical care and improve patient outcomes. The code is available at \href{https://github.com/Wizaaard/X-MultiTask}{https://github.com/Wizaaard/X-MultiTask}.
\end{abstract}

\section{Introduction}
\label{sec:intro}
Evaluating the tailored effects of interventions using observational data is a core challenge in causal inference, with far-reaching implications for decision-making across diverse fields. Estimating treatment effects has long been a central focus in causal inference, epidemiology, and the social sciences \cite{gangl2010causal, wu2024clinical, athey2016recursive}. 
In medicine, particularly in surgical care, this challenge is magnified by high-stakes decisions, heterogeneous patient profiles, and limited access to randomized controlled trials. Understanding how different surgical decisions impact patient outcomes is critical for advancing personalized treatment planning and improving healthcare delivery.

Spinal surgery provides a representative case of these broader challenges. Procedures like spinal fusion are widely used to treat conditions such as degenerative disc disease, scoliosis, and spinal instability \cite{deyo2004spinal}, yet they carry risks of complications that can significantly affect recovery and long-term health \cite{shi2025predicting, devin2015best, pesenti2018risk}. Two key clinical decisions frequently arise in this context: (1) selecting between different surgical approaches (e.g., anterior vs. posterior), and (2) determining whether a patient is likely to benefit from surgery at all. These decisions are often guided by clinical expertise and population-level evidence, but lack robust individualized causal estimates to inform personalized care.

While prior studies have investigated surgical outcomes using descriptive statistics and associational methods \cite{memtsoudis2011increased, qureshi2017comparison}, such approaches do not capture the underlying causal mechanisms that drive differential treatment effects across patients. Recent advances in causal machine learning (ML) provide a powerful framework to rigorously estimate the differential effects of these surgical approaches, addressing key gaps in existing research \cite{leist2022mapping, prosperi2020causal, ben2023adolescent, tamo2024heterogeneous}.

Despite this progress, there remains a critical gap in the application of causal ML to real-world surgical datasets, particularly in modeling complex treatment decisions, integrating high-dimensional clinical features, and addressing challenges like confounding and treatment selection bias. To address this challenge, we develop a multi-task meta-learning framework (X-MultiTask) for treatment effect estimation. Multi-task learning (MTL) is a machine learning paradigm that enables related tasks to be learned simultaneously, allowing the model to leverage shared information across tasks to improve generalization and performance. Using two distinct clinical scenarios in spinal care, surgical approach selection and surgery vs. non-surgical management, we demonstrate how X-MultiTask can inform personalized surgical decision-making, improve outcome prediction, and enhance the interpretability of treatment effects in complex, real-world settings.
The contribution of this work is three-fold:
\begin{itemize}
    \item \textbf{Multi-task meta-learning framework}: X-MultiTask benefits from shared representations across treatment tasks and incorporates inverse probability weighting (IPW) into the loss function to balance treatment assignment, improve robustness, and generalization.
    \item \textbf{Advancing causal inference in surgery}: Provide a rigorous, causally-informed comparison of treatment approaches to spinal fusion surgery.
    \item \textbf{Personalized Treatment Decision-Making}: Estimating treatment effects empowers clinicians to make personalized decisions and improve surgical planning and patient outcomes.
\end{itemize}

\begin{figure*}[t]
    \centering
    \includegraphics[width=\linewidth]{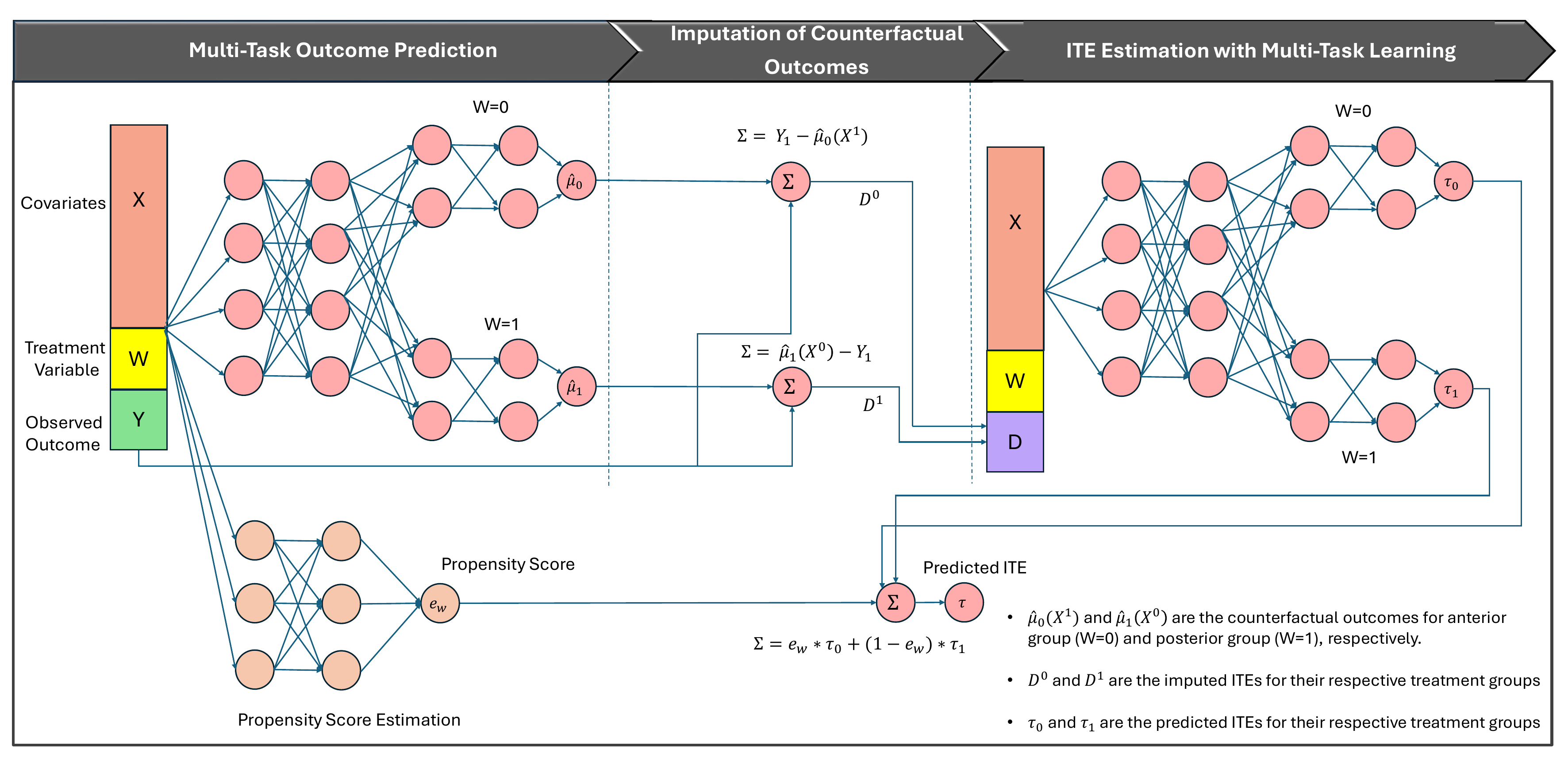}
    \caption{Overview of our Multi-Task Meta-Learning Framework for estimating treatment effects. The framework consists of three key steps: (1) Multi-Task Outcome Prediction, where an MTL model predicts potential outcomes for both treatment groups using shared and task-specific components; (2) Imputation of Counterfactual Outcomes, where counterfactual outcomes are estimated to compute individual treatment effects (ITEs); and (3) ITE Estimation with Multi-Task Learning, where a second MTL model refines treatment effect estimation using shared representations. 
    }

    \label{fig:overview}
\end{figure*}
\section{Related Works}
\label{sec:related_works}
\subsection{Comparative Analyses of Spinal Fusion Surgical Approaches}
Population-based studies have identified distinct trends and risks associated with anterior spinal fusion (ASF), posterior spinal fusion (PSF), and combined anterior/posterior spine fusion (APSF). These findings have been generated through a variety of statistical methods, including bivariate analysis, logistic regression modeling, descriptive statistics, analysis of variance (ANOVA), and multivariate regression analysis \cite{memtsoudis2011perioperative, shamji2009impact, humphreys2001comparison, qureshi2017comparison, vavruch2019surgical, carreon2006treatment}. Memtsoudis et al \cite{memtsoudis2011perioperative} in their study using the National Inpatient Sample, showed that while ASF and APSF are typically performed on younger and healthier patients, APSF carries higher complication and mortality risks.  Similar findings have been reported for cervical spine surgeries, where posterior cervical fusion (PCDF) demonstrated higher rates of complications, including pulmonary and circulatory issues, compared to anterior cervical discectomy and fusion (ACDF) \cite{memtsoudis2011increased}. Furthermore, Shamji et al \cite{shamji2009impact}, in their study on multilevel cervical fusions highlighted that anterior approaches tend to result in fewer perioperative complications and lower resource utilization, whereas posterior techniques are associated with increased morbidity, longer hospital stays, and higher costs.

Literature on comparative analyses of specific surgical techniques reveals trade-offs between anterior and posterior approaches for both cervical and lumbar spine conditions. Research has shown that anterior lumbar interbody fusion (ALIF) is associated with higher complication rates and costs when treating lumbar degenerative disease. However, it is effective for addressing anterior translational instability. In contrast, posterior techniques like transforaminal lumbar interbody fusion (TLIF) results in reduced blood loss and fewer complications, especially in cases of multi-level fusions \cite{humphreys2001comparison, qureshi2017comparison}. In adolescent idiopathic scoliosis, posterior fusion has been shown to achieve superior Cobb angle correction, whereas anterior fusion involves shorter surgeries and fewer instrumented vertebrae, with no significant differences in postoperative alignment or patient-reported outcomes \cite{vavruch2019surgical}. For anterior cervical pseudoarthrosis, posterior fusion has demonstrated superior long-term outcomes in terms of fusion success and revision rates despite greater perioperative challenges \cite{carreon2006treatment}.

\subsection{Causal Inference for Surgical Interventions}

Significant advancements in causal inference and machine learning, grounded in decades of established methodological foundations, has transformed our ability to estimate treatment effects within healthcare. Traditional approaches such as propensity score matching \cite{caliendo2008some}, inverse probability weighting \cite{seaman2013review}, and instrumental variables \cite{angrist1996identification} have proven effective for estimating causal effects from observational data, particularly in binary treatment settings. These methods, along with fundamental contributions to causal inference \cite{rubin2005causal, robins1994correcting}, established the potential outcomes framework that underlies modern approaches. However, the increasing complexity of healthcare interventions has revealed limitations in these traditional methods, particularly when analyzing multi-valued treatments. In such scenarios, increased dimensionality and data sparsity can significantly impact the performance and scalability of conventional approaches. Modern machine learning techniques \cite{Shen_Ma_Vemuri_Castro_Caraballo_Simon_2021b,xu2025medagentgym} have emerged to address these challenges, with methods like causal forests \cite{athey2019estimating}, Bayesian causal inference \cite{hill2011bayesian} and counterfactual neural network \cite{alaa2017deep} offering more flexible, data-driven solutions. 

Causal inference techniques are increasingly used to address complex questions in surgical research. Zhu et al. \cite{zhu2023reasoning} applied a deep learning-based framework with counterfactual predictions and inverse probability weighting (IPW) to estimate individualized treatment effects in low-grade glioma resection. Krishnamoorthy et al. \cite{krishnamoorthy2020causal} demonstrated mediation analysis and joint effects modeling to explore mechanisms underlying perioperative outcomes. Zhang et al. \cite{zhang2022causal} employed marginal structural models with IPW to account for time-varying confounders in longitudinal analyses of laparoscopic surgery.

In light of these developments, this research seeks to bridge methodological innovation with clinical applications in spinal surgery. By introducing a multi-task meta-learning framework, it leverages shared representations across treatment tasks to enhance the robustness and generalization of causal models, particularly in multi-valued treatment settings. 
Through the integration of these advanced methods, the study seeks to empower personalized treatment decision-making, improving surgical planning methods and potentially enhancing patient outcomes, while advancing the application of causal inference in complex, high-dimensional healthcare interventions.

\section{Methodology}
\label{sec:method}

\subsection{Problem Formulation}
We adopt Rubin's potential outcomes model \cite{rosenbaum1983central,rubin2005causal} to quantify the causal impact of surgical approaches on postoperative complication severity in spinal fusion surgery. This framework defines the causal effect for an individual as the difference between their outcome under the factual (observed) and counterfactual (unobserved) conditions. The treatment variable corresponds to the surgical approach, which is binary by definition: $W_i =0$ indicates that patient $i$ underwent the anterior surgical approach (control group), and $W_i =1$ indicates that patient $i$ underwent the posterior surgical approach (treatment group).

Let $Y_i(1)$ and $Y_i(0)$ represent the potential outcomes for the individual $i$ under the posterior and anterior surgical approaches, respectively. The individual treatment effect (ITE) for a patient $i$ with covariates $X_i = x$  is then defined as the expected difference between these two potential outcomes:
\begin{equation}
\tau(x) = \mathbb{E} [Y_i(1) - Y_i(0) \mid X_i = x].
\end{equation}

In cases where the treatment variable is multi-valued—such as comparing multiple surgical techniques or variations within anterior/posterior approaches—the problem can be generalized to compare outcomes across multiple treatment conditions rather than a simple binary treatment-control framework. Given a dataset $\mathcal{D} = {(X_i, W_i, Y_i)}_{i=1}^n$, where:
\begin{itemize}
\item $X_i \in \mathbb{R}^d$ represents the covariates (features) of individual $i$,
\item $W_i \in \{1, 2, \dots, K\}$ denotes the treatment assignment for individual $i$, where $K$ is the total number of treatments (e.g., posterior, anterior, or potential subcategories within each approach),
\item $Y_i \in \mathbb{R}$ is the observed outcome, reflecting the severity of postoperative complications for individual $i$.
\end{itemize}

The primary goal of this work is to estimate the ITE and evaluate the causal impact of surgical approaches on postoperative complication severity. 
Specifically, for any surgical approach $w$ relative to a baseline $w_0$ (e.g., anterior), we seek to estimate:
\begin{equation}
\tau_w(x) = \mathbb{E}[Y(w) - Y(w_0) \mid X = x],
\label{eq:tau_x}
\end{equation}

\subsection{Multi-task Meta-learner}
We propose a multi-task meta-learning framework for estimating heterogeneous treatment effects (HTEs) in observational settings. Our approach leverages the strengths of multi-task learning to simultaneously predict potential outcomes for multiple treatment groups, combining shared and treatment-specific components. The framework consists of three key steps, designed to enhance accuracy, robustness, and interpretability in treatment effect estimation:
\begin{enumerate}
    \item \textbf{Multi-Task Outcome Prediction:}
    
    We begin by fitting an MTL model with two task-specific prediction heads, $\mu_0$(anterior group) and $\mu_1$ (posterior group). These models predict the outcome $Y$ given the covariates $X$ for each treatment group. 
    \item \textbf{Imputation of Counterfactual Outcomes:}

    Using the outcome prediction models from Step 1, we impute the counterfactual outcomes for each individual. For the anterior group ($W_i=0$), we predict what their outcome would have been if they had undergone posterior surgery and subtract their actual outcome. Similarly, for the posterior group ($W_i=1$), we predict what their outcome would have been if they had undergone anterior surgery and subtract their actual outcome. The ITEs for each group are calculated as follows:
    \begin{equation}
        \begin{aligned}
          \begin{array}{ll}
            D_i^1 = Y_i^1 - \hat{\mu}_0(X_i^1), \\
            D_j^0 = \hat{\mu}_1(X_j^0) - Y_j^0. 
          \end{array}
        \end{aligned}
    \end{equation}
    These imputed differences serve as the basis for estimating individual-level treatment effects.
    \item \textbf{ITE Estimation with Multi-Task Learning:}
    
    A second MTL model is trained using the imputed treatment effects from Step 2. This model has two prediction heads, $\tau_0$ and $\tau_1$, designed to estimate the ITEs for each treatment arm while leveraging shared representations to capture common structures across the treatment groups.

\end{enumerate}

The ITE estimate is the weighted average of $\tau_1(x)$ and $\tau_0(x)$ defined as:
\begin{equation}
    \tau(x) = e(x)\tau_0(x) + (1 - e(x))\tau_1(x),
\end{equation}
where $e \in [0,1]$ is the propensity score.

To address treatment assignment imbalance, we incorporate inverse probability weighting (IPW) into the loss function. This ensures robust estimation even when certain treatment groups are underrepresented, by re-weighting samples according to the probability of their treatment assignment.

\textbf{Propensity Score Estimation:}  
We estimate the propensity score \( e_w(X) = P(W = w \mid X) \) for each treatment \( w \) using a neural network.  

\textbf{Weighting Observations:}  
For each treatment \( w \), we apply weights during training:  
\begin{equation}
\text{Weight}_i = \frac{1}{e_w(X_i)} \quad \text{if } W_i = w.
\end{equation}  
This re-weights the loss function for each task in the MTL framework:  
\begin{equation}
\mathcal{L}_w^{\text{IPW}} = \sum_{i: W_i = w} \frac{1}{e_w(X_i)} \cdot \mathcal{L}_w(f_w(X_i), Y_i),
\end{equation}  
where \( f_w \) is the outcome prediction model for treatment \( w \), and \( \mathcal{L}_w \) is the task-specific loss.

\textbf{Joint Training:}  
We also combine the re-weighted losses for all tasks:  
\begin{equation}
\mathcal{L}_{\text{total}} = \sum_{w=1}^K \mathcal{L}_w^{\text{IPW}} + \lambda \mathcal{R}(\theta),
\end{equation}  
where \( \lambda \mathcal{R}(\theta) \) is a regularization term to avoid overfitting and encourage parameter sharing.

\subsection{Assumptions}
To estimate treatment effects, we typically use a conditional approach where the treatment effect is estimated as a function of covariates X in Eq. (\ref{eq:tau_x}).

We then have the following assumptions:
\begin{itemize}
    \item \textbf{Ignorability/Unconfoundedness:} After controlling for the observed covariates, there are no unmeasured confounders that influence both the treatment assignment and the outcomes.
    
    \begin{equation}
        \{Y (1), Y (2), ..., Y (K)\} \perp W \mid X.
    \end{equation}
    Given the covariates, this assumption posits that the potential outcomes are conditionally independent of the treatment assignment. 
    \item \textbf{Positivity:} Each unit has a non-zero chance of receiving either treatment. This ensures that we have a mix of treated and untreated units across the levels of covariates, where
    \begin{equation}
        0 < P(W_i) < 1.
    \end{equation}
    \item \textbf{Stable Unit Treatment Value Assumption (SUTVA):} Given any two units $i$ and $j$, the potential outcomes of $i$, are unaffected by the treatment assignment of unit $j$,
    \begin{equation}
        \{Y_i(1), Y_i(2), ..., Y_i (K)\} \perp W_j , \forall j \neq i. 
    \end{equation}
    
\end{itemize}

\subsection{Evaluation}
Causal inference is inherently challenging due to the fundamental problem of missing counterfactuals—each unit is observed under only one treatment, making true Individual Treatment Effects (ITEs) unobservable. To evaluate our model, we use both outcome prediction metrics and proxy-based treatment effect metrics.

\begin{enumerate}
    \item \textbf{Outcome Prediction Evaluation}

    We assess classification performance using the Area Under the ROC Curve (AUC-ROC), which quantifies the model’s ability to distinguish outcomes (e.g. complication severity or PROs) across treatment groups.
    \item \textbf{Treatment Effect Evaluation}

    We evaluate treatment effect accuracy using the Precision in Estimation of Heterogeneous Effects (PEHE) and Absolute Error in Average Treatment Effect (ATE) \cite{tran2023data}. Since true effects are not directly observable, we approximate them using Nearest Neighbor (NN) matching based on covariate similarity.
\end{enumerate}

Nearest Neighbor Precision in Estimation of Heterogeneous Effects (NN-PEHE) assumes that the nearest control and treated neighbor of each unit ($\text{NN}_{0}(i)$ and $\text{NN}_{1}(i)$) provide a good estimation of the treatment effect. Once the units are matched, the substitute ground-truth treatment effect is calculated as 
\begin{equation}
    \hat{\tau}_{NN}(\mathbf{X_{i}}, Y_{i}, W_{i}) = Y_{i}^{\text{NN}}(1) -  Y_{i}^{\text{NN}}(0),
\end{equation}
This value is then used to approximate the mean squared error for $\tilde{\tau}$:
Then the overall error is calculated as:
\begin{equation}
        \begin{aligned}
          \begin{array}{ll}
            \hat{\epsilon}_{\text{NN-PEHE}} = \frac{1}{N} \sum_{i = 1}^{n} \Bigl( \hat{\tau}_{NN}(\mathbf{X_{i}}, Y_{i}, W_{i}) - \tilde{\tau}(\mathbf{X}_{i}) \Bigr)^{2}.\\[4pt]
            \hat{\epsilon}_{ATE} =  |\hat{\tau}_{ATE}^{\text{True approx}}  - \tilde{\tau}_{ATE}^{\text{predicted}}|
          \end{array}
        \end{aligned}
    \end{equation}

where $$\hat{\tau}_{ATE}^{\text{True approx}} = \frac{1}{n}\sum_{i = 1}^{n}\hat{\tau}_{NN}(\mathbf{X_{i}}, Y_{i}, W_{i})$$  
$$\tilde{\tau}_{ATE}^{\text{predicted}} = \frac{1}{n}\sum_{i = 1}^{n}\tilde{\tau}(\mathbf{X}_{i})$$

A lower NN-PEHE indicates better matching and greater precision in treatment effect estimation based on nearest neighbors. In contrast, a higher NN-PEHE suggests poor matching or larger discrepancies between the estimated and true effects.

\section{Experimental Results}
\label{sec:results}
We evaluate the proposed X-MultiTask framework in two clinically relevant scenarios:
\begin{itemize} 
    \item  \textbf{RQ1:} What is the causal effect of anterior (ASF) vs. posterior (PSF) spinal fusion on postoperative complication severity, categorized into four ordinal levels (0 to 3)? Where, 0–no complication, 1–one complication, 2–two complications, and 3–three or more complications. 
    \item  \textbf{RQ2:} What is the individual treatment effect of PSF vs. non-surgical management in adolescents with idiopathic scoliosis (AIS), based on patient-reported outcomes (PROs)?
\end{itemize}

\subsection{Datasets}
\subsubsection{Medical Informatics Operating Room Vitals and Events Repository (MOVER) Dataset}
We used the publicly available MOVER \cite{samad2023mover}, specifically the EPIC subset, encompassing records from 39,685 patients and a total of 64,354 surgeries (2017-2022). 
We identify a subset of preoperative and postoperative follow-up data for 1,228 spinal fusion surgeries with 10,354 variables.
The preprocessing pipeline builds a structured, feature-rich dataset of spinal fusion patients by filtering diagnosis, procedure, and coding data for surgical relevance. It integrates pre-op labs (471 tests), medications (3319 entries), diagnosis history (5994 terms), procedure events (9 types), and LDA data (28 flow, 43 sites), all timestamped to retain only pre-surgical records. Features are standardized, aggregated, and pivoted into wide format, and a capped complication count is added as the outcome label, yielding a clean cohort for supervised learning.
A comprehensive summary of the dataset features is provided in Table \ref{tab:dataset}. The code is available at \href{https://github.com/Wizaaard/X-MultiTask}{https://github.com/Wizaaard/X-MultiTask}

\subsubsection{Private AIS Dataset}
We utilized electronic health records from 368 adolescent idiopathic scoliosis (AIS) patients who underwent spinal fusion surgery at Shriners Children's, a multi-institutional pediatric healthcare system with 22 locations across North America. The dataset for this study was drawn specifically from two participating sites: Shriners Children’s Greenville and Shriners Children’s Lexington. These hospitals contribute to the Setting Scoliosis Straight (SSS) initiative, Surgeon Performance Program (SPP), and Quality Improvement Registry, which promote standardized care and outcomes tracking across centers. This cohort includes both preoperative and longitudinal postoperative data, encompassing demographic information, detailed radiographic measurements, and patient-reported outcomes (PROs). PROs were assessed using the validated Scoliosis Research Society-22 Revised (SRS-22R) questionnaire \footnote{https://www.srs.org/professionals/online-education-and-resources/patient-outcome-questionnaires}, capturing domains such as function, pain, self-image, mental health, and satisfaction. Clinical experts extracted radiographic features from full-length standing posterior/anterior and lateral spine radiographs. A comprehensive list of the variables included in the dataset is provided in Table 1. While the cohort is limited to two sites, the standardized data collection protocols and multi-center infrastructure of Shriners Children's support the potential for broader generalizability and future prospective validation.

\begin{table*}[!t]
\centering
\caption{Summary of variables across the MOVER-EPIC and Private AIS datasets, grouped by clinical category. Patient-reported outcomes in the AIS dataset are derived from the SRS-22R questionnaire, scored on a 5-point Likert scale (1 = worst, 5 = best). Radiographic metrics reflect standard spinal alignment measurements. 
}

\label{tab:dataset}
\fontsize{9}{11}\selectfont
\setlength{\tabcolsep}{0.65em}
\resizebox{\linewidth}{!}{
\begin{tabular}{llcc}\\
\toprule
\textbf{Category} & \textbf{Variable} & \textbf{MOVER-EPIC} & \textbf{Private AIS Dataset} \\
\midrule

\multirow{7}{*}{\textbf{Demographics}} 
& Age & 60 $\pm$ 16 years & 12 $\pm$ 2 years \\
& Gender & 57.6\% Male & 79.1\% Female \\
& Race & -- & 74.8\% White \\
& Weight & 180 $\pm$ 46 lb & -- \\
& Height & 5 ft 7 in $\pm$ 2 in & -- \\
& ASA Status & Most frequent: Severe systemic disease & -- \\
& Comorbidities & -- & 74\% With Comorbidities \\

\midrule
\textbf{Surgery-Related} 
& Anesthesia Type & Most frequent: General & -- \\
& Discharge Disposition & Most frequent: Home routine & -- \\

\midrule
\textbf{Medications}
& Pre-operative Medications & 3319 entries & -- \\

\midrule
\textbf{History}
& Diagnosis History & 5994 unique diagnoses & -- \\

\midrule
\textbf{Laboratory Results}
& Pre-op Lab Test Values & 471 entries & -- \\
& Pre-op Abnormal Flags & 471 entries & -- \\

\midrule
\multirow{2}{*}{\textbf{LDA (Lines, Drains, Airways)}}
& Flow Measurements (Pre-op) & 28 entries & -- \\
& Placement Site (Pre-op) & 43 entries & -- \\

\midrule
\textbf{Complications} 
& Complication Type & Most frequent: Anesthesia-related & -- \\

\midrule
\textbf{Procedure Events} 
& Pre-operative Events & 9 unique events & -- \\

\midrule
\multirow{6}{*}{\textbf{Radiographic (Lateral)}} 
& C7–Sacrum Alignment & -- & 12.64 $\pm$ 28.78 \\
& T12–Sacrum Lordosis & -- & 59.47 $\pm$ 12.55 \\
& T10–L2 Segment & -- & 9.71 $\pm$ 9.07 \\
& T2–T12 Kyphosis & -- & 32.21 $\pm$ 13.17 \\
& T2–T5 Segment & -- & 12.64 $\pm$ 28.78 \\
& T5–T12 Segment & -- & 22.74 $\pm$ 13.4 \\

\midrule
\multirow{10}{*}{\textbf{Radiographic(Posterior/Anterior)}} 
& C7 to CSVL Offset & -- & -4.76 $\pm$ 15.62 \\
& Lumbar Curve Magnitude & -- & 44.45 $\pm$ 16 \\
& Lumbar Curve Direction & -- & 94\% Left \\
& Thoracic Curve Magnitude & -- & 59.89 $\pm$ 14.34 \\
& Thoracic Curve Direction & -- & 94\% Right \\
& Upper Thoracic Curve Magnitude & -- & 27.36 $\pm$ 11.12 \\
& Upper Thoracic Curve Direction & -- & 94.85\% Left \\
& T1 Tilt Angle & -- & 7.47 $\pm$ 6.23 \\
& Thoracic Apex to C7 Plumb Line & -- & 38.53 $\pm$ 18.74 \\
& Thoracolumbar/Lumbar Apex to CSVL & -- & 33.65 $\pm$ 26.01 \\

\midrule
\multirow{5}{*}{\textbf{Patient Outcomes (SRS-22R)}} 
& Function (Q5, 9, 12, 15, 18) & -- & 4.35 $\pm$ 0.36\\
& Mental Health (Q3, 7, 13, 16, 20) & -- & 3.96 $\pm$ 0.5\\
& Satisfaction (Q21, 22) & -- &  4.38 $\pm$ 0.53\\

\bottomrule
\end{tabular}
}
\end{table*}

\subsection{Assumption Validity}
While ignorability, positivity, and SUTVA are essential for causal inference, their validity can be questioned in real-world clinical settings. Factors such as surgeon experience, institutional protocols, and clinical judgments may influence surgical choices and outcomes.
To address this, we include a variety of preoperative features, such as diagnosis history, medications, lab results, radiographic findings, and patient-reported outcomes. However, residual confounding is still a concern.
Positivity violations occur when certain subgroups are exclusively assigned to a treatment, like high-risk patients receiving only one surgical approach, due to clinical guidelines or contraindications. To mitigate this, we monitor estimated propensity scores for near-violations and apply reweighting techniques for better overlap.
SUTVA assumes no interference between patients' treatments and outcomes. While this holds in individual surgical interventions, hospital-level factors, like shared resources and care practices, could still lead to potential violations.

\subsection{Baselines}
We evaluate several baseline causal machine learning models to estimate treatment effects and model heterogeneous treatment responses. Below, we provide an overview of each method and its relevance in causal inference.

\begin{itemize}
    \item \textbf{S-Learner}: A single neural network trained on both treated and control data with treatment as a feature. Treatment effects are estimated by contrasting predictions under different treatment values. 
    \item \textbf{T-Learner}: Two separate models are trained for treated and control groups. Treatment effects are estimated as the difference in predictions. Offers flexibility but requires balanced data.
    \item \textbf{X-Learner} \cite{kunzel2019metalearners}: Trains separate models for treated and control groups, imputes counterfactuals, then learns treatment effects from these imputed values, improving accuracy in imbalanced datasets.
    \item \textbf{Counterfactual Regression (CFR)} \cite{johansson2016learning}: Builds on TARNet with an added distribution-balancing loss using Integral Probability Metrics (IPM) to reduce selection bias.
    \item  {\textbf{Deep Counterfactual Networks with Propensity-Dropout (DCN-PD)} \cite{alaa2017deep}: Models outcomes via multitask learning with shared and outcome-specific layers, using propensity-dependent dropout to address selection bias.}
\end{itemize}

\subsection{Implementation}
We implement all models using PyTorch and optimize them with the AdamW optimizer. Model training is conducted on H100 GPU for high-performance computation. To streamline configurations, we store all data settings and hyperparameters in structured configuration files.

To ensure a fair and objective evaluation, we perform an extensive hyperparameter search using AUC score as the primary evaluation metric for outcome prediction. To enhance efficiency, we leverage Optuna, a powerful hyperparameter optimization library, to accelerate the tuning process. The search space for all hyperparameters is detailed in Table \ref{tab:hyperparameters}.

\begin{table}[ht]
    \centering
    \caption{Hyperparameter Search Space for Model Optimization: The range of values explored for various hyperparameters during the optimization process, conducted using Bayesian search with the Optuna library to efficiently identify optimal configurations.}
    \begin{tabular}{ll} \\
        \toprule
        \textbf{Range} & \textbf{Functionality} \\
        \midrule
        \{64, 128, ..., 8192\} & Hidden layer size \\
        $[0.1, 0.7]$ & Dropout probability \\
        $[1e-5, 1e-1]$ & Learning rate \\
        \{32, 64, 128, 256\} & Batch size \\
        \{2, 3, ..., 8\} & Number of shared layers \\
        \{1, 2, ..., 5\} & Task-specific layers \\
        $[1e-5, 1e-3]$ & Regularization term \\
        \bottomrule
    \end{tabular}
    \label{tab:hyperparameters}
\end{table}

\subsection{Model Evaluation}
We evaluate our outcome prediction framework using two datasets within surgical decision-making contexts. For RQ1, we address a multi-class classification task, predicting postoperative complication severity (ranging from level 0 to 3), based on preoperative EHR data from patients undergoing anterior (control) or posterior (treated) surgical approaches. To enable fine-grained treatment effect estimation across severity levels, we decompose this task into a series of binary classification problems, one per severity class.
For RQ2, we focus on predicting post-treatment PROs in adolescents with AIS, comparing those who underwent PSF to those who received no surgical intervention. We consider three PRO domains: Function, Mental Health, and Satisfaction. Each domain score represents the average of patient-reported Likert-scale responses (0 to 5) across questions within that domain.

\begin{table*}[ht]
    \centering
    \caption{AUC scores and their 95\% confidence intervals across different models for two datasets: MOVER-EPIC and Private AIS.}
    \resizebox{\linewidth}{!}{
    \begin{tabular}{llcccc|lccc}\\
        \toprule
        &\multicolumn{5}{c|}{\textbf{MOVER-EPIC Dataset (Complication Severity Levels)}} & \multicolumn{4}{c}{\textbf{Private AIS Dataset (PRO Domains)}} \\
        \cmidrule(lr){2-6} \cmidrule(lr){7-10}
        \textbf{Model} & \textbf{Group} & \textbf{Severity 0} & \textbf{Severity 1} & \textbf{Severity 2} & \textbf{Severity 3}
        & \textbf{Group} & \textbf{Function} & \textbf{Mental Health} & \textbf{Satisfaction} \\
        \midrule
        \multirow{2}{*}{S-Learner} 
        & Anterior  & $0.94 \pm 0.08$ & $0.85 \pm 0.08$ & $0.83 \pm 0.09$ & $0.84 \pm 0.15$ 
        &  {No Surgery}  &    {$0.76 \pm 0.14$} &  {$0.80 \pm 0.14$} &  {$0.60 \pm 0.37$} \\
        & Posterior & $0.72 \pm 0.22$ & $0.79 \pm 0.07$ & $0.74 \pm 0.10$ & $0.83 \pm 0.07$ 
        &  {PSF} &  {$0.86 \pm 0.15$} &  {$0.68 \pm 0.20$} &  {$0.68 \pm 0.18$} \\
        \midrule
        \multirow{2}{*}{T-Learner} 
        & Anterior  & $0.86 \pm 0.16$ & $0.75 \pm 0.11$ & $0.67 \pm 0.13$ & $0.80 \pm 0.14$ 
        &  {No Surgery}  &  {$0.73 \pm 0.23$}  &  {$0.87 \pm 0.11$} &  {$0.91 \pm 0.16$} \\
        & Posterior & $0.78 \pm 0.18$ & $0.80 \pm 0.05$ & $0.71 \pm 0.08$ & $0.84 \pm 0.06$ 
        &  {PSF} &  {$0.87 \pm 0.17$} &   {$0.76 \pm 0.13$} &  {$0.73 \pm 0.16$} \\
        \midrule
        \multirow{2}{*}{X-Learner} 
        & Anterior  & $0.91 \pm 0.13$ & $0.79 \pm 0.10$ & $0.67 \pm 0.14$ & $0.79 \pm 0.13$ 
        &  {No Surgery}  &  {$0.74 \pm 0.19$} &  {$0.90 \pm 0.09$} &  {$0.67 \pm 0.35$} \\
        & Posterior & $0.75 \pm 0.16$ & $0.77 \pm 0.06$ & $0.69 \pm 0.09$ & $0.83 \pm 0.06$ 
        &  {PSF} &  {$0.56 \pm 0.32$} &   {$0.63 \pm 0.18$} &  {$0.82 \pm 0.13$} \\
        \midrule
        \multirow{2}{*}{CFR} 
        & Anterior  & $0.24 \pm 0.17$ & $0.88 \pm 0.15$ & $0.77 \pm 0.22$ & $0.84 \pm 0.17$ 
        &  {No Surgery}  &  {$0.74 \pm 0.06$} &  {$0.83 \pm 0.12$} &  {$0.57 \pm 0.12$}  \\
        & Posterior & $0.78 \pm 0.14$ & $0.79 \pm 0.06$ & $0.81 \pm 0.06$ & $0.84 \pm 0.06$ 
        &  {PSF} &  {$0.38 \pm 0.08$} &  {$0.59 \pm 0.18$} &  {$0.89 \pm 0.04$} \\
        \midrule
        \multirow{2}{*}{ {DCN-PD}} 
        &  {Anterior}  &  {$0.67 \pm 0.18$} &  {$0.66 \pm 0.12$} &  {$0.39 \pm 0.16$} &  {$0.60 \pm 0.18$} 
        &  {No Surgery}  &  {$0.69 \pm 0.23$} &  {$0.86 \pm 0.13$} &  {$0.92 \pm 0.13$}  \\
        &  {Posterior} &  {$0.63 \pm 0.30$} &  {$0.34 \pm 0.08$} &  {$0.47 \pm 0.12$} &  {$0.59 \pm 0.10$} 
        &  {PSF} &  {$0.61 \pm 0.16$} &  {$0.69 \pm 0.16$} &  {$0.77 \pm 0.20$} \\
        \midrule
        \multirow{2}{*}{X-MultiTask} 
        & Anterior  & $0.95 \pm 0.06$ & $0.80 \pm 0.10$ & $0.74 \pm 0.14$ & $0.87 \pm 0.11$ 
        &  {No Surgery}  &  {$0.82 \pm 0.16$} &  {$0.87 \pm 0.11$} &  {$0.98 \pm 0.04$} \\
        & Posterior & $0.70 \pm 0.21$ & $0.78 \pm 0.07$ & $0.76 \pm 0.10$ & $0.83 \pm 0.07$ 
        &  {PSF} &  {$0.80 \pm 0.11$} &  {$0.73 \pm 0.15$} &  {$0.76 \pm 0.15$} \\
        \bottomrule
    \end{tabular}}
    \label{tab:auc_scores}
\end{table*}

Table~\ref{tab:auc_scores} reports AUC scores (with 95\% confidence intervals) for each model across both the MOVER-EPIC and Private AIS datasets. The S-Learner performs well in the anterior group (severity 0: 0.94, severity 1: 0.85) but drops significantly in the posterior group (severity 0: 0.72), indicating poor generalizability.  {It shows modest performance in AIS PROs, with stronger Function prediction in PSF (0.86) than in no-surgery (0.76).} The T-Learner shows more balanced performance (severity 0 AUCs: 0.86 anterior, 0.78 posterior) but underperforms in severity 2 (anterior: 0.67). The X-Learner achieves high anterior AUC (severity 0: 0.91) but drops in posterior severity 0 (0.75) and severity 2 (0.69), suggesting difficulty with posterior low-severity cases. CFR is highly variable, with very low in anterior severity 0 (0.24) but better in posterior severity 2 (0.81) and severity 3 (0.84), indicating potential overfitting or severity imbalance issues. The X-MultiTask model consistently outperforms others, showing strong performance across anterior (severity 0: 0.95; severity 3: 0.87) and posterior groups (severity 3: 0.83; severity 2: 0.76).  {In the AIS cohort, it delivers robust and balanced predictions for both PSF (Function: 0.80; Satisfaction: 0.76) and no-surgery groups (Function: 0.82; Satisfaction: 0.98), highlighting its superior generalizability and stability across treatment arms and outcome types.}

Table~\ref{tab:pehe} compares treatment effect estimation using NN-PEHE and ATE Error across the MOVER-EPIC and Private AIS datasets. On MOVER-EPIC, X-MultiTask achieves the best individualized treatment effect estimation (NN-PEHE overall: 0.2778), slightly outperforming DCN-PD (0.2783), S-Learner (0.2987), and X-Learner (0.3065), while CFR performs worst (2.9877), particularly in severity 1 (4.09) and severity 2 (3.42). In ATE Error, X-Learner (0.0579) and X-MultiTask (0.0763) lead, with S-Learner showing moderate performance (0.1331) but higher error in severity 1 (0.2362) and severity 3 (0.2316); CFR again performs worst (0.5672).
On the Private AIS dataset, X-MultiTask also yields the lowest overall NN-PEHE (0.2551), outperforming all baselines in Function (0.1959) and Satisfaction (0.1899). For ATE estimation, it achieves the best overall score (0.0902), with consistently low error across all PRO domains. In contrast, S-Learner and X-Learner perform reasonably (ATE: 0.1909 and 0.1328, respectively), while T-Learner and DCN-PD show less stability across domains.
Overall, X-MultiTask demonstrates the most robust and accurate treatment effect estimation across both individual and population levels, generalizing well across complication severity levels and PRO outcomes.

\begin{table*}[ht]
    \centering
    \caption{Comparison of treatment effect estimation methods using $\hat{\epsilon}_{\text{NN-PEHE}}$ and $\hat{\epsilon}_{ATE}$ across MOVER-EPIC and Private AIS. Lower values indicate better estimation accuracy: $\hat{\epsilon}_{\text{NN-PEHE}}$ evaluates individual-level precision, and $\hat{\epsilon}_{ATE}$ measures population-level error.}
    \resizebox{\linewidth}{!}{
    \begin{tabular}{lcccccc|lcccc}\\
        \toprule
        & \multicolumn{5}{c}{\textbf{MOVER-EPIC Dataset} — $\hat{\epsilon}_{\text{NN-PEHE}}$} & & \multicolumn{4}{c}{\textbf{Private AIS Dataset} — $\hat{\epsilon}_{\text{NN-PEHE}}$} \\
        \cmidrule(lr){2-6} \cmidrule(lr){8-11}
        \textbf{Model} & severity 0 & severity 1 & severity 2 & severity 3 & Overall & & Function & Mental Health & Satisfaction & Overall\\
        \midrule
        S-Learner   & 0.1160 & 0.4638 & 0.3203 & 0.2946 & 0.2987 & &  {0.2024} &  {0.5148} &  {0.1971} &  {0.4572}\\
        T-Learner   & 0.6691 & 2.2196 & 1.4350 & 1.7083 & 1.5080 & &  {0.2332} &  {0.4893} &  {0.3291} &  {0.3505}\\
        X-Learner   & 0.1128 & 0.4725 & 0.3628 & 0.2778 & 0.3065 & &  {0.2262} &  {0.3907} &  {0.1938} &  {0.2702}\\
        CFR         & 2.8950 & 4.0929 & 3.4200 & 1.5428 & 2.9877 & &  {0.5306} &  {0.8749} &  {0.3231} &  {0.5762} \\
         {DCN-PD} &  {0.1138} &  {0.4276} &  {0.3144} &  {0.2572} &  {0.2783} & &  {0.3403} &  {0.3487} &  {0.2217} &  {0.3036}\\
        X-MultiTask & 0.1126 & 0.4285 & 0.3157 & 0.2539 & 0.2778 & &  {0.1959} &  {0.3795} &  {0.1899} &  {0.2551}\\
        \midrule
        & \multicolumn{5}{c}{\textbf{MOVER-EPIC Dataset} — $\hat{\epsilon}_{ATE}$} & & \multicolumn{4}{c}{\textbf{Private AIS Dataset} — $\hat{\epsilon}_{ATE}$} \\
        \cmidrule(lr){2-6} \cmidrule(lr){8-11}
        \textbf{Model} & severity 0 & severity 1 & severity 2 & severity 3 & Overall & & Function & Mental Health & Satisfaction & Overall\\
        \midrule
        S-Learner   & 0.0032 & 0.2362 & 0.0614 & 0.2316 & 0.1331 & &  {0.1103} &  {0.3684} &  {0.0939} &  {0.1909}\\
        T-Learner   & 0.2324 & 0.4449 & 0.0162 & 0.6653 & 0.3397 & &  {0.0759} &  {0.0718} &  {0.1997} &  {0.1158}\\
        X-Learner   & 0.0201 & 0.0575 & 0.0583 & 0.0958 & 0.0579 & &  {0.1080} &  {0.1962} &  {0.0942} &  {0.1328} \\
        CFR         & 0.2217 & 0.9107 & 0.1819 & 0.9548 & 0.5672 & &  {0.2893} &  {0.3524} &  {0.2243} &  {0.2887} \\
         {DCN-PD} &  {0.0316} &  {0.0628} &  {0.1138} &  {0.1231} &  {0.0828} & &  {0.0923} &  {0.0882} &  {0.1098} &  {0.0968} \\
        X-MultiTask & 0.0173 & 0.0483 & 0.0871 & 0.1527 & 0.0763 & &  {0.0494} &  {0.1697} &  {0.0516} &  {0.0902} \\
        \bottomrule
    \end{tabular}}
    \label{tab:pehe}
\end{table*}

The ablation study presented in Table~\ref{tab:ablation} demonstrates the contribution of each component in the \texttt{X-MultiTask} model architecture. The full model achieves strong predictive performance (Avg AUC: 0.849 for control, 0.775 for treated) and competitive treatment effect estimation accuracy (NN-PEHE: $0.278 \pm 0.044$, ATE error: $0.075 \pm 0.048$). Notably, removing the shared representation layers (“No Shared Layers”) leads to the most substantial drop in performance across all metrics, particularly in control AUC (0.721) and PEHE (0.359), underscoring the critical role of multi-task shared feature learning. This architectural component enables the model to generalize across treatment groups and tasks by capturing shared structure in a way that single-headed or decoupled models (e.g., traditional T-Learner or X-Learner) cannot.

Removing inverse propensity weighting (“No IPW”) slightly degrades treatment effect estimation, reflected by increases in both PEHE and ATE errors. This highlights the utility of reweighting to mitigate treatment assignment bias, even when the imbalance is modest. Interestingly, replacing learned propensity scores with a constant (0.5) results in comparable performance, suggesting that in this dataset, where treatment assignment is not highly skewed, explicit estimation of propensity scores may offer limited additional benefit. Removing regularization improves ATE error slightly but increases PEHE, indicating mild overfitting in heterogeneous effect estimation.

Compared to related methods, \texttt{X-MultiTask} offers a distinct contribution by combining multi-task learning and representation sharing for counterfactual estimation in a unified, end-to-end trainable architecture. Unlike CFR and DCN-PD, which focus on balancing representations through domain adversarial loss or stochastic dropout informed by propensity scores, our approach leverages multi-task learning to jointly optimize control and treated outcome prediction with shared feature representations. Moreover, while X-Learner decomposes treatment effect estimation into a two-step imputation process, it does not benefit from shared inductive biases across tasks. The performance gains observed in the ablation study empirically validate the importance of this architectural integration, supporting \texttt{X-MultiTask} as a meaningful advancement tailored to real-world surgical datasets where sample sizes are limited and treatment heterogeneity is high.

\begin{table*}[ht]
    \centering
    \caption{
    Ablation study results for the \texttt{X-MultiTask} model on the MOVER-EPIC dataset. 
    We report average AUC across complication severity (95\% confidence interval) for both control and treated groups, and treatment effect estimation metrics.
    }
    \begin{tabular}{lcccc}\\
    \toprule
     {\textbf{Model Variant}} &  {\textbf{Avg AUC (Control)}} &  {\textbf{Avg AUC (Treated)}} &  {\textbf{Avg $\hat{\epsilon}_{NN-PEHE}$}} &  {\textbf{Avg $\hat{\epsilon}_{ATE}$}} \\
    \midrule
     {X-MultiTask} &  {0.849 $\pm$ 0.087} &  {0.775 $\pm$ 0.081} &  {0.278 $\pm$ 0.044} &  {0.075 $\pm$ 0.048} \\
     {No IPW} &  {0.831 $\pm$ 0.105} &  {0.775 $\pm$ 0.076} &  {0.283 $\pm$ 0.046} &  {0.079 $\pm$ 0.053} \\
     {Constant Propensity} &  {0.862 $\pm$ 0.095} &  {0.765 $\pm$ 0.088} &  {0.280 $\pm$ 0.045} &  {0.081 $\pm$ 0.052} \\
     {No Regularization} &  {0.843 $\pm$ 0.101} &  {0.782 $\pm$ 0.092} &  {0.285 $\pm$ 0.047} &  {0.065 $\pm$ 0.046} \\
     {No Shared Layers} &  {0.721 $\pm$ 0.167} &  {0.744 $\pm$ 0.111} &  {0.359 $\pm$ 0.061} &  {0.065 $\pm$ 0.052} \\
    \bottomrule
    \end{tabular}
    \label{tab:ablation}
\end{table*}

\subsection{RQ1: Causal effect of anterior versus posterior spinal fusion approaches on the severity of postoperative complications}

\subsubsection{Subgroups Analysis: Clustering and Average Treatment Effect Analysis}

\begin{figure}[t]
    \centering
    \includegraphics[width=\linewidth]{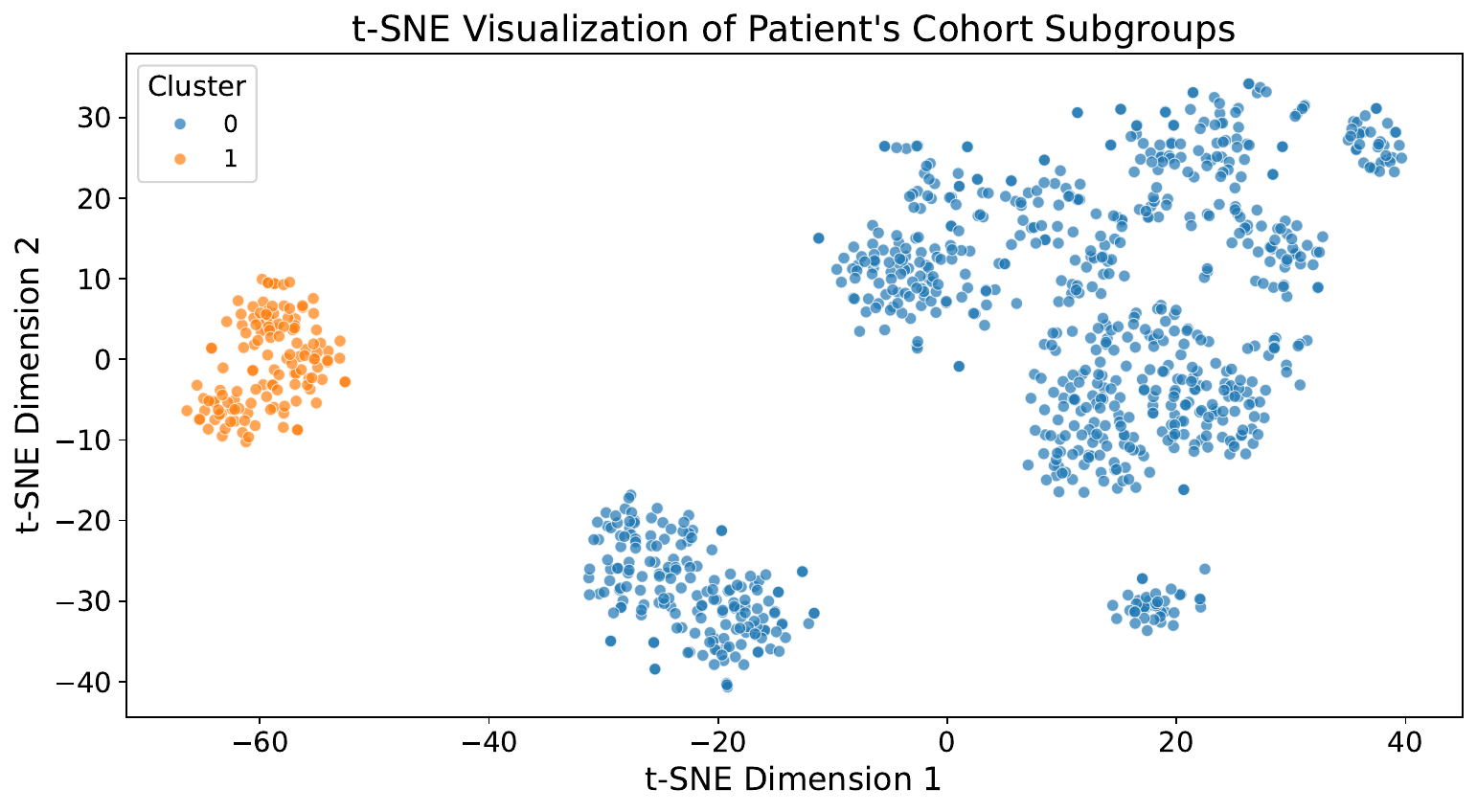}
    \caption{t-SNE plot of patient subgroups based on preoperative EHR, derived through unsupervised clustering analysis. This achieves a Silhouette score of 0.8304 indicating a well-defined clustering structure, revealing two primary subgroups.}
    \label{fig:subgroups}
\end{figure}

\begin{table*}[ht]
    \centering
    \caption{
    Numeric features identified as significantly different between subgroups using the KS test (p-value $<$ 0.01).}
    \begin{tabular}{lccc}\\
        \toprule
        \textbf{Numeric Feature} & \textbf{P-Value} & \textbf{Cluster 0 (Mean ± Std)} & \textbf{Cluster 1 (Mean ± Std)} \\
        \midrule
        SARS coronavirus 2 RNA\_value & 1.0216e-06 & 0.2677 ± 0.4428 & 0.5149 ± 0.4998 \\
        Age & 4.83247e-04 & 0.5786 ± 0.2140 & 0.6467 ± 0.2158 \\
        Creatinine\_value & 2.272743e-03 & 0.0702 ± 0.0730 & 0.0597 ± 0.0291 \\
        Hemoglobin\_value & 4.608868e-03 & 0.0271 ± 0.1624 & 0.0224 ± 0.1479 \\
        Specific gravity\_value & 4.894967e-03 & 0.0189 ± 0.1361 & 0.0149 ± 0.1213 \\
        Erythrocytes\_value & 6.348054e-03 & 0.0153 ± 0.1229 & 0.0299 ± 0.1702 \\
        \bottomrule
    \end{tabular}
    
    \label{tab:ks_test_features}
\end{table*}

\begin{figure}[t]
    \centering
    \includegraphics[width=\linewidth]{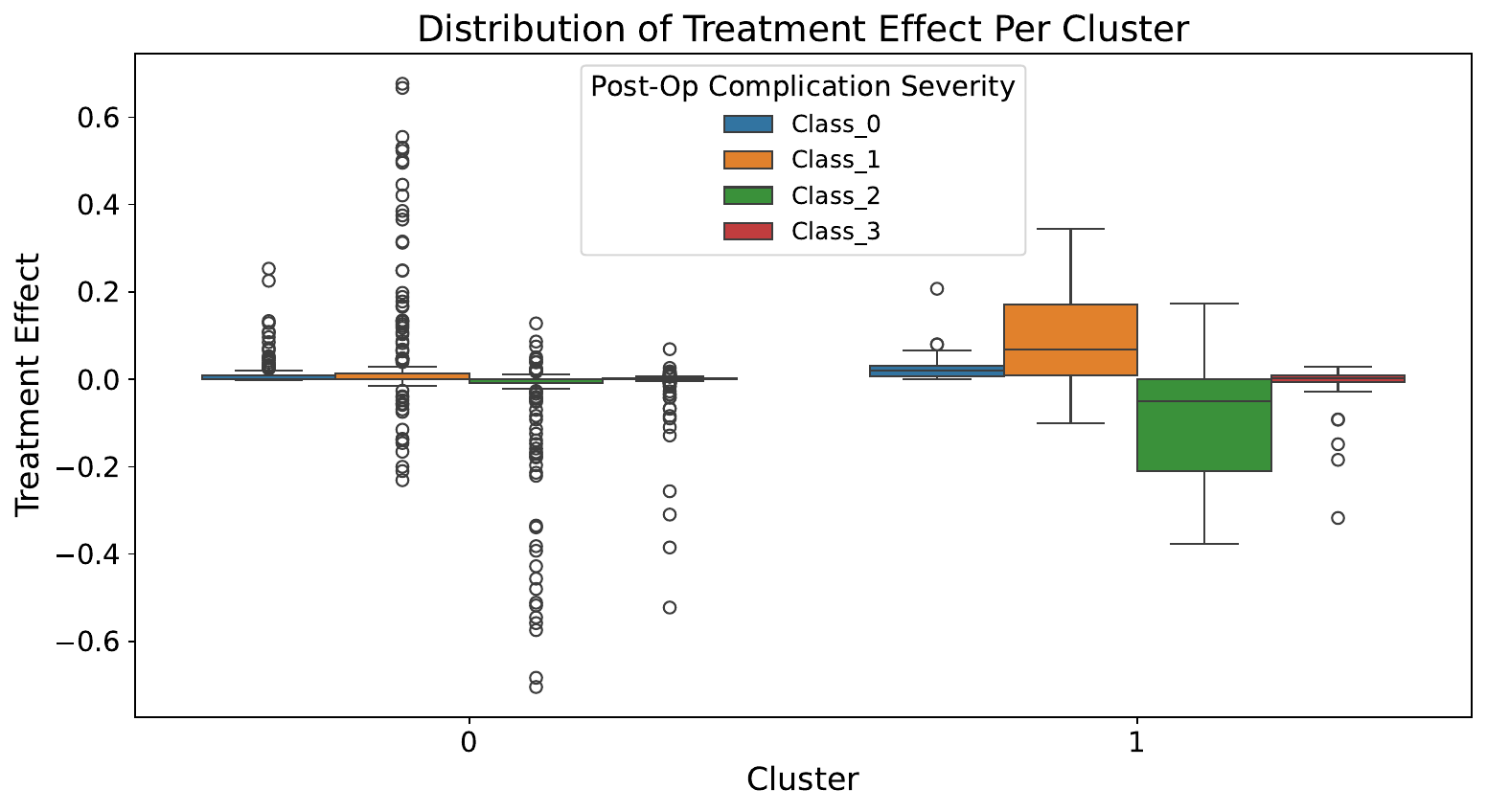}
    \caption{ Distribution of Treatment Effects Across Clusters and Severity classes: The box plot visualizes the distribution of treatment effects for different postoperative complication severity classes within two distinct subgroups (0 and 1).}
    \label{fig:effect_subgroups}
\end{figure}

Using unsupervised clustering, we identified two distinct patient subgroups based on clinical and lab profiles, supported by a high Silhouette score (0.8304), indicating well-separated subgroups with differing clinical characteristics and treatment responses (Figure \ref{fig:subgroups}). To assess statistical significance, numerical features were compared using the Kolmogorov-Smirnov test ($p<0.01$), revealing significant inter-cluster differences in SARS-CoV-2 RNA levels, age, creatinine, hemoglobin, specific gravity, and erythrocytes (Table \ref{tab:ks_test_features}). Categorical variables, analyzed via the Chi-square test ($p<0.01$), showed strong associations with subgroup membership, including comorbidities, medication history, and mobility impairments (Table \ref{tab:condensed_ks_test}). Subgroup 1 had a higher mean SARS-CoV-2 RNA value ($0.5149 \pm 0.4998$ vs. $0.2677 \pm 0.4428$), suggesting greater infection burden, which can elevate complication risk. Age and creatinine were also higher in subgroup 1, reflecting increased frailty and impaired renal function, both known surgical risk factors. Hemoglobin and erythrocyte levels were lower in subgroup 1, pointing to possible anemia and reduced oxygen-carrying capacity, which can impair healing. Specific gravity was slightly higher in subgroup 0, indicating better hydration or renal status. Subgroup 1 also showed higher rates of comorbidities, polypharmacy, and mobility impairments, all of which are associated with elevated surgical risk. Elevated discharge disposition values in subgroup 1 further suggest these patients required more intensive post-treatment care. As shown in Figure \ref{fig:effect_subgroups}, ATE estimates varied across outcome classes, with subgroup 1 exhibiting a more pronounced treatment response, positive or negative, underscoring the need to account for baseline heterogeneity when evaluating intervention effects.

\begin{table}[ht]
    \centering
    \caption{
    categorical features that show significant differences between subgroup 0 and subgroup 1 based on the Chi-squared test, with a significance threshold of p-value $<$ 0.01. }
    \resizebox{\linewidth}{!}{
    \begin{tabular}{lccc}\\
        \toprule
        \multirow{2}{*}{\textbf{Categorical Feature}} & \multirow{2}{*}{\textbf{P-Value}} & \multicolumn{2}{c}{\textbf{\% True}} \\
        \cmidrule(lr){3-4}
         & & \textbf{Subgroup 0} & \textbf{Subgroup 1} \\
        \midrule
        Impaired Mobility History & 1.00e-05 & 0.47\% & 5.22\% \\
        RSV RNA Abnormal & 1.01e-05 & 10.61\% & 24.63\% \\
        Venous Insufficiency History & 1.61e-05 & 0.00\% & 2.99\% \\
        Abnormal Gait History & 1.61e-05 & 0.00\% & 2.99\% \\
        Preoperative Exam History & 8.96e-05 & 0.24\% & 3.73\% \\
        Influenza B RNA Abnormal & 1.13e-04 & 82.55\% & 67.91\% \\
        Influenza A RNA Abnormal & 1.13e-04 & 82.55\% & 67.91\% \\
        Benzodiazepines Abnormal & 2.12e-04 & 89.86\% & 78.36\% \\
        Diabetes Type 1.5 History & 2.32e-04 & 0.12\% & 2.99\% \\
        Lumbar Fusion History & 2.34e-04 & 5.31\% & 14.18\% \\
        Decreased Activity & 4.28e-04 & 0.00\% & 2.24\% \\
        Bilateral Myopia & 4.28e-04 & 0.00\% & 2.24\% \\
        Hypertension History & 4.73e-04 & 1.18\% & 5.97\% \\
        Ethanol Abnormal & 4.95e-04 & 87.74\% & 76.12\% \\
        Gait Impairment History & 7.37e-04 & 0.94\% & 5.22\% \\
        Iron Deficiency Anemia & 1.08e-03 & 0.71\% & 4.48\% \\
        Methadone Abnormal & 1.11e-03 & 87.74\% & 76.87\% \\
        Amphetamine Abnormal & 1.88e-03 & 89.03\% & 79.19\% \\
        Benzoylecgonine History & 3.67e-03 & 87.97\% & 78.36\% \\
        THC Abnormal & 3.90e-03 & 89.62\% & 80.60\% \\
        \bottomrule
    \end{tabular}}
    
    \label{tab:condensed_ks_test}
\end{table}

\begin{figure*}[!t]
    \centering
    \begin{subfigure}
        \centering
        \includegraphics[width = 0.49\linewidth]{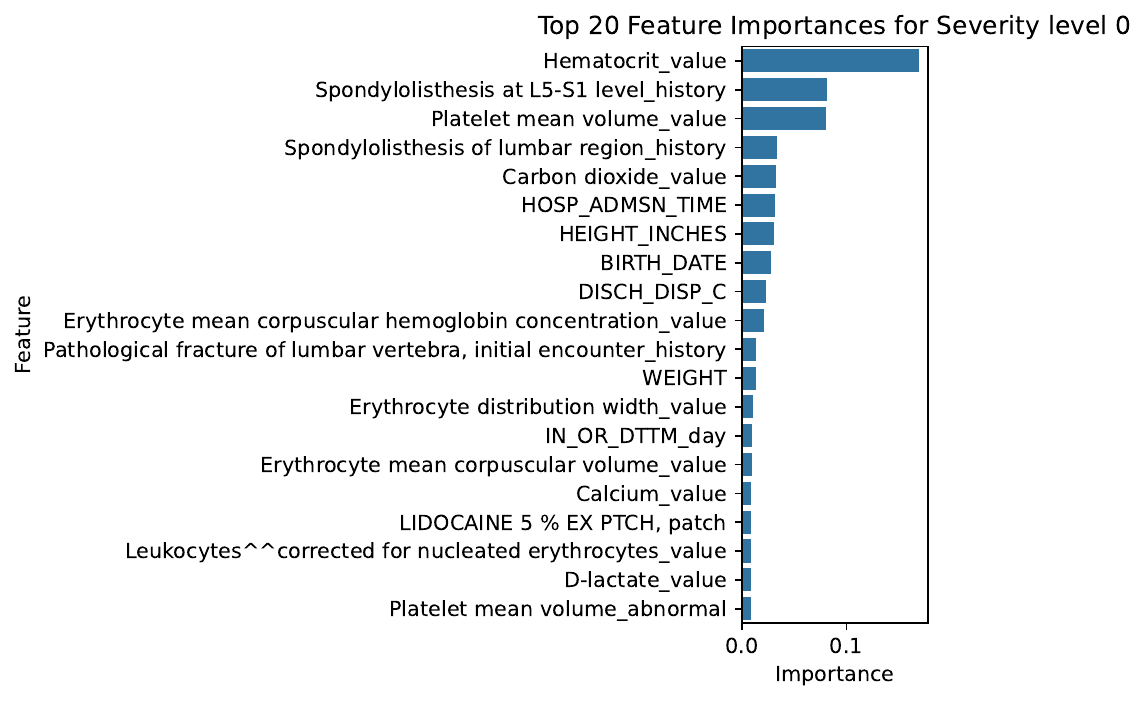}
    \end{subfigure}
    \hfill
    \begin{subfigure}
        \centering
        \includegraphics[width = 0.49\linewidth]{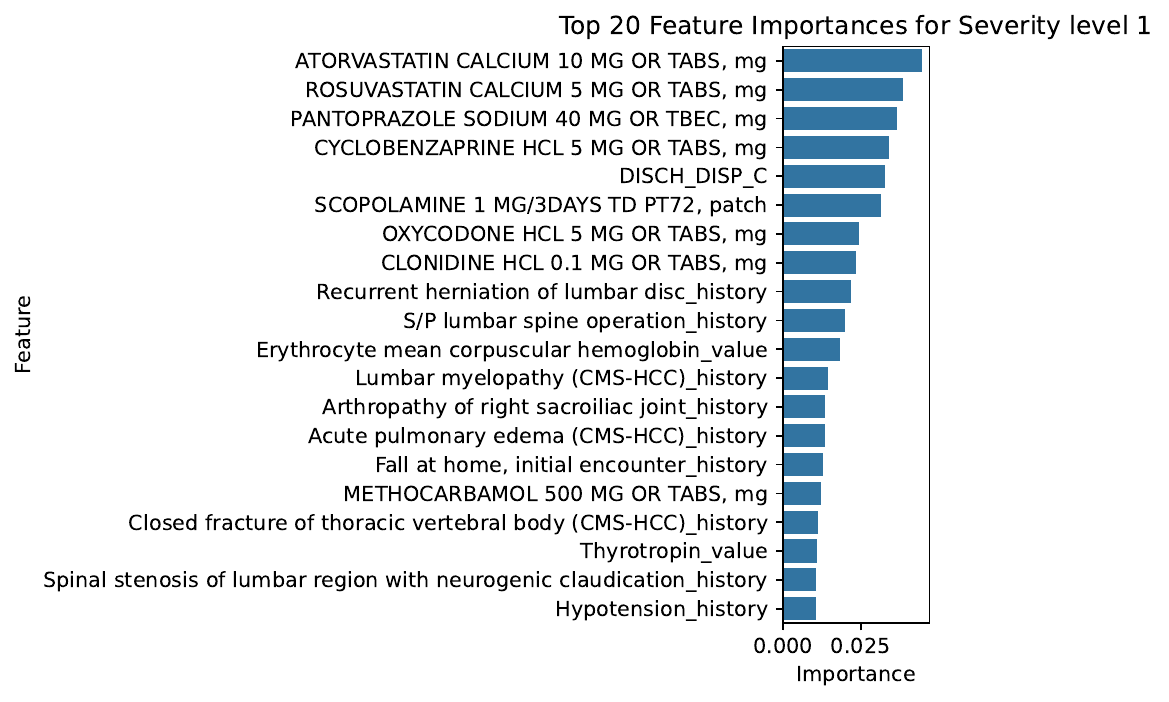}
    \end{subfigure}
    
    \vskip\baselineskip  
    
    \begin{subfigure}
        \centering
        \includegraphics[width = 0.49\linewidth]{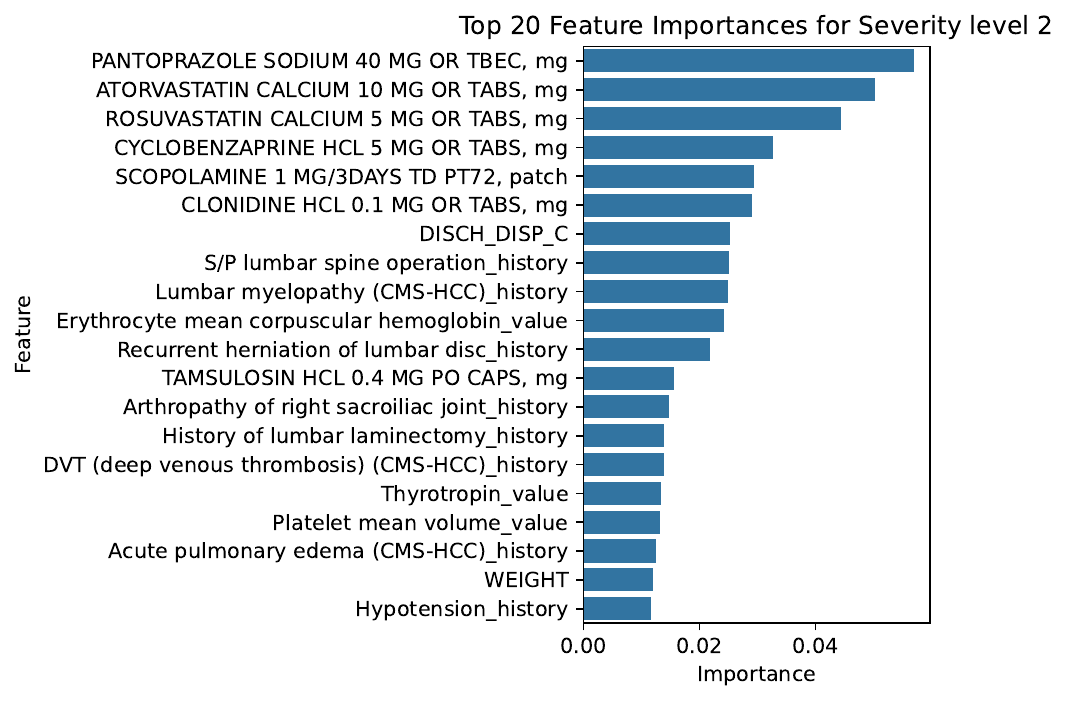}
    \end{subfigure}
    \hfill
    \begin{subfigure}
        \centering
        \includegraphics[width = 0.49\linewidth]{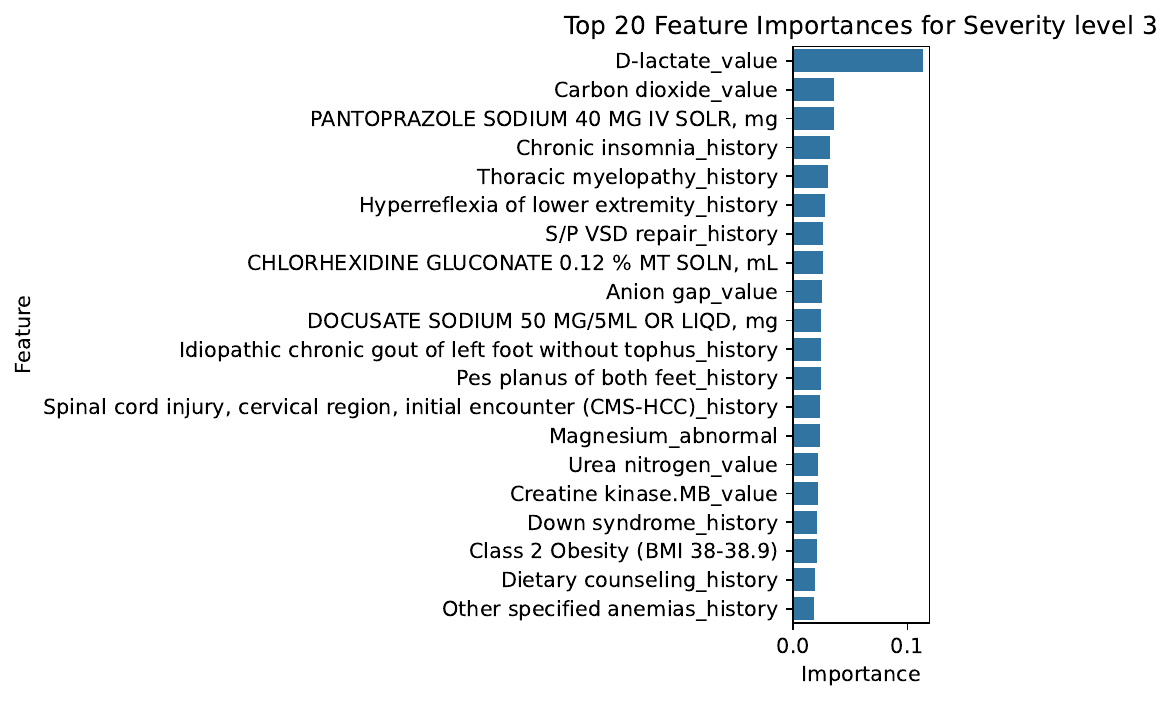}
    \end{subfigure}

    \caption{Top features influencing treatment effect prediction across complication severity levels (0–3). Severity 0 is driven by hematologic and musculoskeletal factors; severity 1 by medication history and spinal conditions. Severity 2 reflects pharmacologic, musculoskeletal, and metabolic interplay, while severity 3 is dominated by metabolic, respiratory, and spinal cord-related indicators.}
    \label{fig:mainzz}
\end{figure*}
\subsubsection{Individual Treatment Effect Analysis}
To evaluate how treatment effects vary across clinically distinct subgroups, we trained separate XGBoost regressors for each complication severity within the two clusters identified through unsupervised analysis. We then examined the top 20 features per model to understand which clinical and laboratory variables most influenced treatment response (Figure \ref{fig:mainzz}). This stratified modeling approach enables fine-grained interpretation of treatment heterogeneity across patient profiles.

In subgroup 0, the lower-risk cohort, predictive features for patients with no complications (class 0) included laboratory values such as anion gap, leukocyte count, platelet count, weight, and age. These results indicate that physiological and immune-related variables had relatively higher importance in predicting favorable treatment response. In class 1 (one complication), feature importance shifted toward erythrocyte indices and comorbidity indicators (e.g., atorvastatin use, lumbar spinal stenosis). For higher severity levels (class 2 and 3), features such as PANTOPRAZOLE SODIUM 40 MG, thyroid function markers, CRP, and D-lactate appeared among the top-ranked, suggesting that systemic and inflammatory factors became more relevant as complication severity increased.

In subgroup 1, the higher-risk cohort, even patients without complications showed influence from cervical stenosis, potassium, calcium, and abnormal urobilinogen; suggesting a background of renal-metabolic and neurologic vulnerability. Among patients with severity level 1 complications, dominant features included markers of systemic dysfunction such as erythrocyte indices, blood urea nitrogen, and coagulation parameters. Severity level 2 was primarily associated with abnormalities in hemoglobin, carbon dioxide, phosphate, and hematocrit, suggesting impairments in oxygen transport and acid-base balance. In severity 3, the most severe cases were shaped by acetaminophen use, ionized calcium, and platelet levels, reflecting medication burden, metabolic instability, and potential procedural risk factors.

When a new patient presents, their preoperative data (labs, history, medications, etc.) can be used to assign them to a clinical cluster and then compute ITE across complication classes. Overall, ITEs stratified by complication severity help identify which surgical approach is safer or riskier for a given patient, enabling personalized, risk-adjusted surgical planning and better patient engagement.

\subsection{RQ2: Treatment effect of undergoing PSF versus non-surgical management in AIS population}

\begin{figure*}[!t]
    \centering
    \begin{subfigure}
        \centering
        \includegraphics[width = 0.3\linewidth]{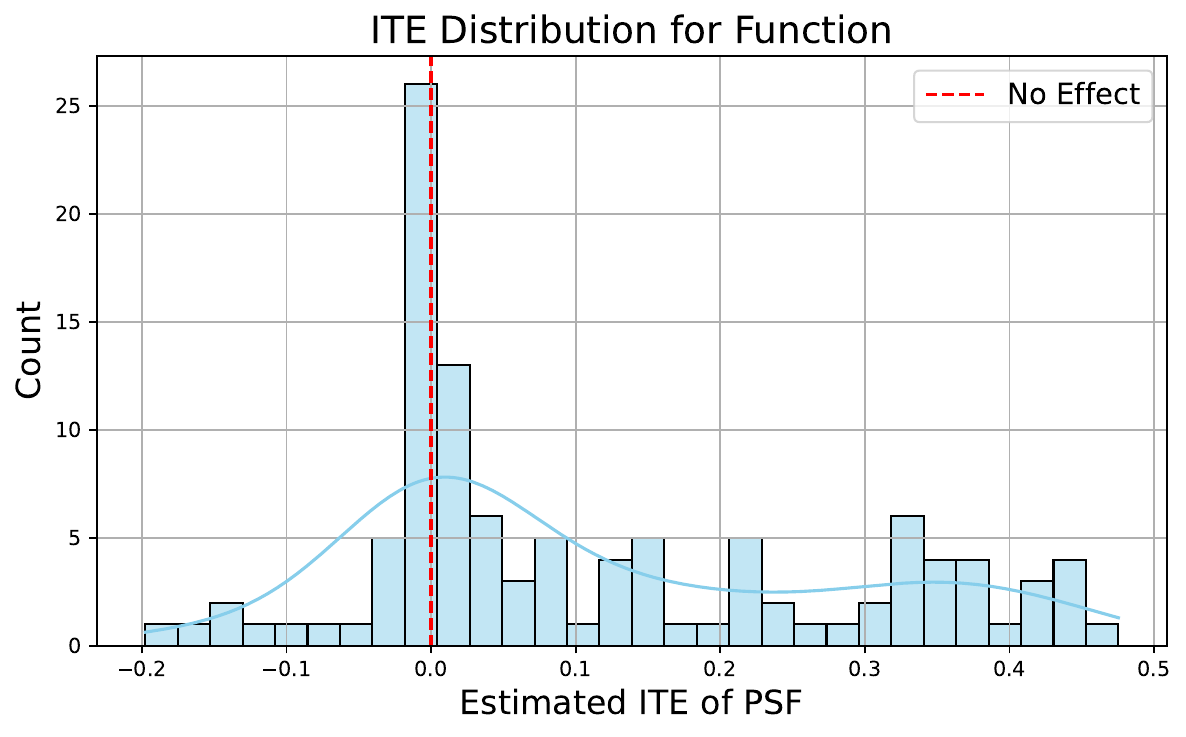}
    \end{subfigure}
    \hfill
    \begin{subfigure}
        \centering
        \includegraphics[width = 0.3\linewidth]{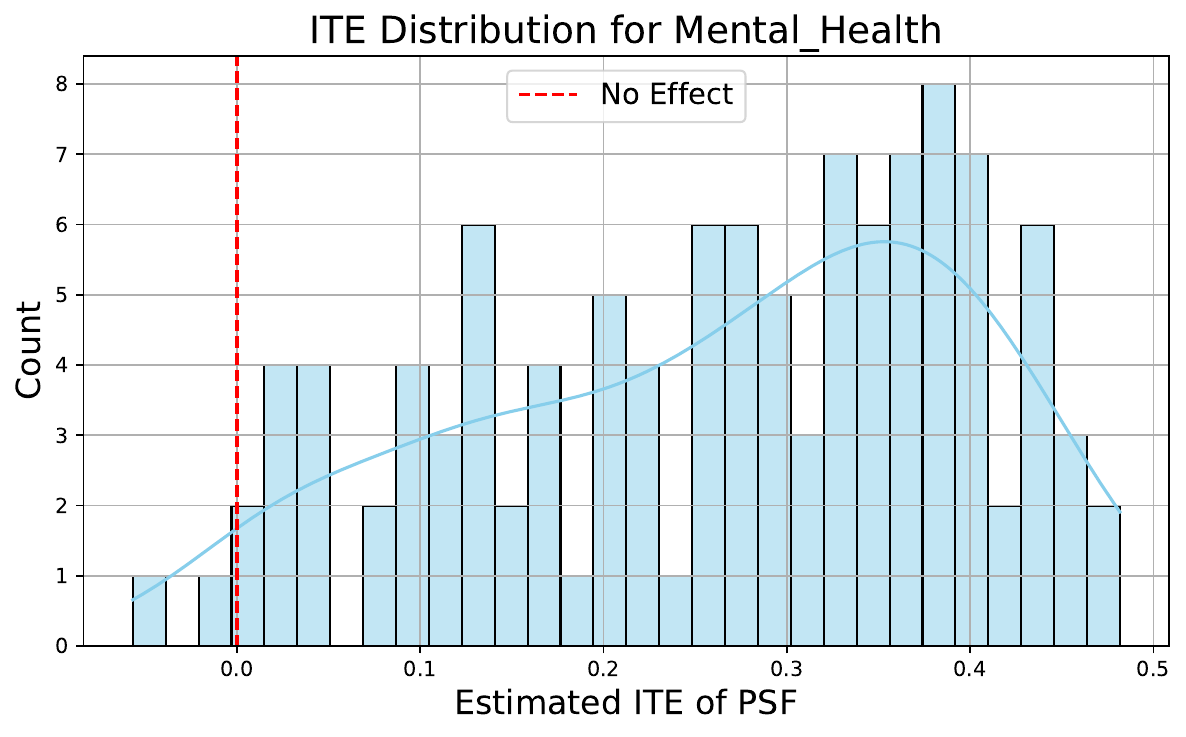}
    \end{subfigure}
    \hfill
    \begin{subfigure}
        \centering
        \includegraphics[width = 0.3\linewidth]{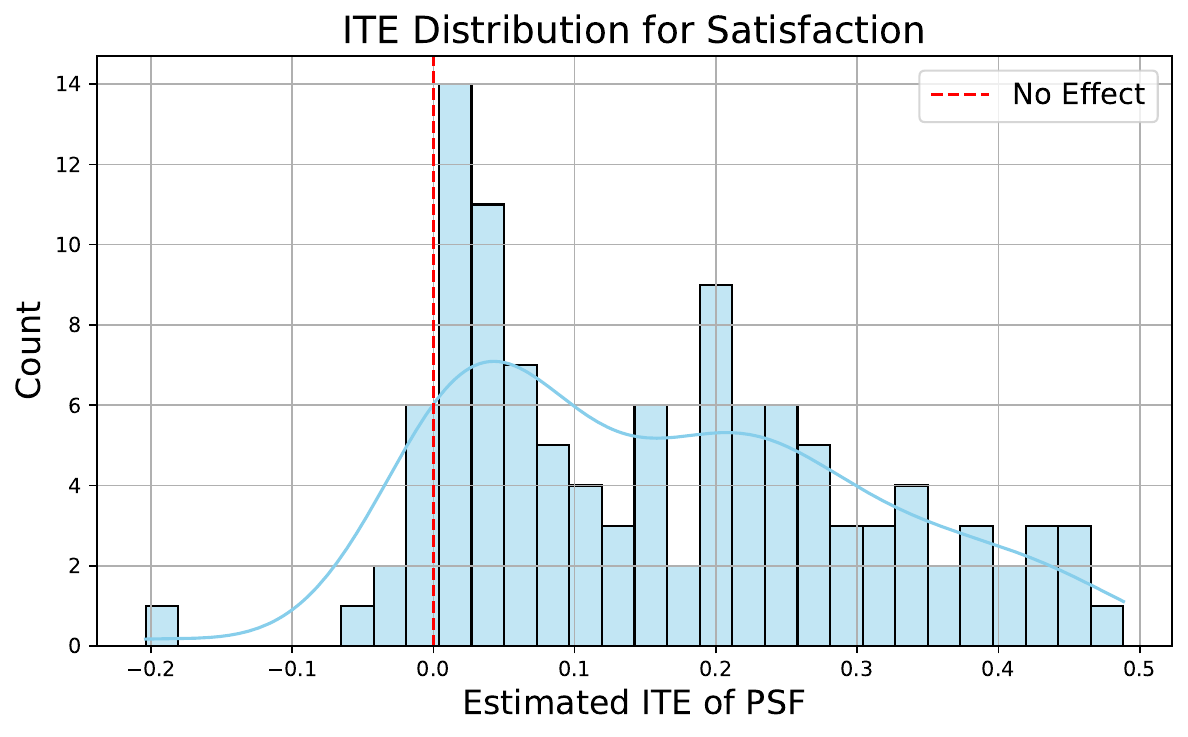}
    \end{subfigure}

    \caption{Distribution of predicted ITEs for PSF across three PRO domains in the AIS cohort. Positive ITE values indicate improved outcomes with PSF, while negative values suggest better outcomes with non-surgical management. The distributions reveal treatment effect heterogeneity across patients, supporting personalized surgical decision-making.}
    \label{fig:ais_ite_distribution}
\end{figure*}

The estimated ITEs across the three PRO domains (Figure~\ref{fig:ais_ite_distribution}) suggest notable heterogeneity in predicted responses to posterior spinal fusion (PSF) among AIS patients. In the Function domain, the average ITE is modest (0.1157), with a broad distribution (range: –0.20 to 0.48), and approximately 71.4\% of patients predicted to benefit. This indicates the potential for differential functional outcomes post-surgery, and may motivate further investigation into predictive markers of benefit. By contrast, the Mental Health domain shows a higher and more uniform predicted benefit (mean ITE: 0.2619), with 97.3\% of patients predicted to improve, suggesting stronger model confidence in this domain. Similarly, the Satisfaction domain shows generally positive predictions (mean ITE: 0.1626; 94.6\% improved). These results provide preliminary evidence that mental health and satisfaction outcomes may be more consistently improved by PSF than functional outcomes, though further validation is needed. While these model-based predictions are not definitive clinical evidence, they highlight the potential utility of individualized treatment effect modeling in supporting future surgical decision-making frameworks for AIS.
\section{Conclusion}
\label{sec:conclusion}
In this study, we developed and validated a multi-task meta-learning framework that leverages causal inference to personalize surgical decision-making. 
Our results demonstrate that by modeling distinct clinical choices as related tasks, the X-MultiTask framework can learn robust, shared representations from complex clinical data to deliver superior outcome predictions compared to existing methods. 
Applying this approach to spinal fusion surgery, we successfully identified patient-specific factors that modify the risks associated with anterior and posterior surgical approaches. Furthermore, in the AIS cohort, our model quantified the heterogeneous benefits of surgery on patient-reported outcomes, revealing that the expected improvements in function and mental health vary significantly among individuals.
X-MultiTask has the potential to transform clinical practice by empowering surgeons and patients to collaboratively select the treatment strategy that best aligns with the patient's specific clinical profile and personal preferences.

\section*{Acknowledgements}
This research was supported by Shriners Children's Hospital and the Georgia Institute of Technology through the Forecasting Unexpected Signal Change in Posterior Spinal Fusion (FUSION) Project. Additional support was provided in part by the AI Makerspace of the College of Engineering and other research cyberinfrastructure resources and services offered by the Partnership for an Advanced Computing Environment (PACE) at the Georgia Institute of Technology, Atlanta, Georgia, USA.

\bibliography{ref}
\bibliographystyle{icml2025}


\end{document}